# Using Causal Information and Local Measures to Learn Bayesian Networks

Wai Lam *
Department of Computer Science
University of Waterloo
Waterloo, Ontario,
Canada, N2L 3G1

Fahiem Bacchus †
Department of Computer Science
University of Waterloo
Waterloo, Ontario,
Canada, N2L 3G1

## Abstract

In previous work we developed a method of learning Bayesian Network models from raw data. This method relies on the well known minimal description length (MDL) principle. The MDL principle is particularly well suited to this task as it allows us to tradeoff, in a principled way, the accuracy of the learned network against its practical usefulness. In this paper we present some new results that have arisen from our work. In particular, we present a new *local* way of computing the description length. This allows us to make significant improvements in our search algorithm. In addition, we modify our algorithm so that it can take into account partial domain information that might be provided by a domain expert. The local computation of description length also opens the door for local refinement of an existent network. The feasibility of our approach is demonstrated by experiments involving networks of a practical size.

## 1 Introduction

Bayesian networks, advanced by Pearl [Pea86], have become an important paradigm for representing and reasoning under uncertainty. Systems based on Bayesian networks have been constructed in a number of different application areas, ranging from medical diagnosis [BBS91], to oil price reasoning [Abr91]. Despite these successes, a major obstacle to using Bayesian networks lies in the difficulty of constructing them in complex domains: there is a knowledge engineering bottleneck. Clearly, it would be extremely useful if the construction process could be fully or partly automated. A useful approach, that has recently being pursued by a number of authors, is to attempt to build, or learn, a network model from raw data. In practice, raw data is often available from databases of records.

We have developed a new approach to learning Bayesian network models [LB93b]. Our approach is based on Rissanen's Minimal Description Length (MDL) [Ris78] principle. The MDL principle offers a means for trading off model complexity and accuracy, and our experiments have demonstrated its suitability for this task. In this paper we present some significant improvements to our original system [LB93b] which (1) make it more efficient, (2) allow it to take into consideration domain information about causation and ordering, and (3) allow local refinement of an existing network.

These improvements are mainly based on a new analysis of the description length parameter that shows how we can evaluate the description length of a proposed network via *local* computations involving only a node and its parents. This localized evaluation of description length allows us to develop an improved searching mechanism that performs well even in fairly large domains. In addition, it allows us to modify our search procedure so that it can take into consideration domain knowledge of *direct causes* as well as *partial orderings* among the variables. Such partial information about the structure of the domain is quite common and in many cases it can reduce the complexity of the searching process during learning.

The localized evaluation of description length also allows us to modify an existing Bayesian network by refining a local part of it. By refining the network we obtain a more accurate model, or adapt an existing model to an environment that has changed over time.

In the sequel we will first describe, briefly, the key features of our previous work, concentrating in particular on the advantages of the MDL approach. Then we derive a new localized version of the description length computation. Using this we develop an algorithm that searches for a good network model, taking into consideration causal and ordering information about the do main. Finally, we discuss the results of various experiments we have run that demonstrate the effectiveness of our approach. The experimental results of our work on local refinement of an existing network are not yet complete, but we will close with a brief discussion of the method. The experiment results will be reported in our full report [LB93a].

*Wai Lam's work was supported by an OGS scholarship. His e-mail address is wlam1@math.uwaterloo.ca

†Fahiem Bacchus's work was supported by NSERC and by IRIS. His e-mail address is fbacchus@logos.uwaterloo.ca

## 2  Learning Bayesian Networks

Much early work on learning Bayesian networks shares the common disadvantage of relying on assumptions about the underlying distribution being learned. For example, Chow and Liu [CL68] developed methods that construct tree structured networks; hence their method provides no guarantees about the accuracy of the learned structure if the underlying distribution cannot be expressed by a tree structure. The approach of Rebane and Pearl [RP87], as well as that of Geiger et. al. [GPP90], suffers from the same criticism, except that they are able to construct singly connected networks. Sprites et al.[SS90] as well as Verma and Pearl [VP90, PV91] develop approaches that are able to construct multiply connected networks, but they require the underlying distribution to be *dag-isomorphic*.[1]

The problem with making an assumption about the underlying distribution is that generally we do not have sufficient information to test our assumption. The underlying distribution is unknown; all we have is a collection of records in the form of variable instantiations. Hence, in practice these methods offer no guarantees about the accuracy of the learned model except in the rare circumstances where we know something about the underlying distribution.

Our approach can construct an accurate model from an unrestricted range of underlying distributions, and it is capable of constructing networks of arbitrary topology, i.e., it can construct multiply connected networks. The ability to construct a multiply connected networks is sometimes essential if the network is to be a sufficiently accurate model of the underlying distribution.

Although multiply connected networks allow us to more accurately model the underlying distribution they have computational as well as conceptual disadvantages. Exact belief updating procedures are, in the worst case, computationally intractable over multiply connected networks [Coo90]. Moreover, even if an approximation algorithm is used, e.g., the stochastic simulation methods of [CC90, Pea87, SP90], highly connected networks still require the storage and estimation of an exponential number of conditional probability parameters.[2] Hence, even if a highly connected network is more accurate, in practice it might not be *as useful* a model as a simpler albeit slightly less accurate model. In addition to the computational disadvantages the causal relationships between the variables are conceptually more difficult to understand in a complex network.

Hence, we are faced with a tradeoff. More complex networks allow for more accurate models, but at the same time such models may be of less practical use than simpler models. The MDL principle allows us to balance this tradeoff: our method will learn a less complex network if that network is sufficiently accurate, and at the same time it is still capable of learning a complex network if no simpler one is sufficiently accurate. This seems to be a particularly appropriate approach to take in light of the fact that we only have a sample of data points from the underlying distribution. That is, it seems inappropriate to try to learn the "most accurate" model of the underlying distribution given that the raw data only provides us with an approximate picture of it.

Among other works on learning Bayesian networks, the most closely related is that of Cooper and Herskovits [CH91]. They use a Bayesian approach that, like ours, is capable of learning multiply connected networks. However, as with all Bayesian approaches they must choose some prior distribution over the space of possible networks. One way of viewing the MDL principle is as a mechanism for choosing a reasonable prior that is biased towards simpler models. Cooper and Herskovits [CH91] investigate a number of different priors, but it is unclear how any particular choice will influence the end result. The MDL principle, on the other hand, allows the system designer (who can choose different ways of encoding the network) to choose a prior based on principles of computational efficiency. For example, if we prefer to learn networks in which no node has more than 5 parents, we can choose an encoding scheme that imposes a high penalty on networks that violate this constraint.

### 2.1  Applying the MDL Principle

The MDL principle is based on the idea that the best model representing a collection of data items is the model that minimizes the sum of

1. the length of the encoding of the model, and
2. the length of the encoding of the data *given the model*,

both of which can be measured in bits. A detailed description of the MDL principle with numerous examples of its application can be found in [Ris89].

To apply the MDL principle to the task of learning Bayesian networks we need to specify how we can perform the two encodings, the network itself (item 1) and the raw data given a network (item 2).

---

[1] A distribution is dag-isomorphic if there is some dag that displays all of its dependencies and independencies [Pea88].

[2] The number of parameters required is exponential in the maximum number of parents of node.

**Encoding the Network**   Our encoding scheme for the networks has the property that the higher the topological complexity of the network the longer will be its encoding. To represent the structure of a Bayesian network we need for each node a list of its parents and a list of its conditional probability parameters.

Suppose there are $n$ nodes in the problem domain. For a node with $k$ parents, we need $k \log_2(n)$ bits to list its parents. To represent the conditional probabilities, the encoding length will be the product of the number of bits required to store the numerical value of each conditional probability and the total number of conditional probabilities that are required. In a Bayesian network, a conditional probability is needed for every distinct instantiation of the parent nodes and node itself (except that one of these conditional probabilities can be computed from the others due to the fact that they all sum to 1). For example, if a node that can take on 4 distinct values has 2 parents each of which can take on 3 distinct values, we will need $3^2 \times (4-1)$ conditional probabilities.

Hence, the total description length for a particular network will be:

$$\sum_{i=1}^{n}[k_i \log_2(n) + d(s_i - 1) \prod_{j \in F_i} s_j], \qquad (1)$$

where there are $n$ nodes; for node $i$, $k_i$ is the number of its parent nodes, $s_i$ is the number of values it can take on, and $F_i$ is the set of its parents; and $d$ represents the number of bits required to store a numerical value. For a particular problem domain, $n$ and $d$ will be constants. This is not the only encoding scheme possible, but it is simple and it performs well in our experiments.

By looking at this equation, we see that highly connected networks require longer encodings. First, for many nodes the list of parents will become larger, and second the list of conditional probabilities we need to store for that node will also increase. In addition, networks in which nodes that have a larger number of values have parents with a large number of values will require longer encodings. Hence, the MDL principle, which is trying to minimize the sum of the encoding lengths, will tend to favor networks in which the nodes have a smaller number of parents (i.e., networks that are less connected) and also networks in which nodes taking on a large number of values are not parents of nodes that also take on a large number of values.

In Bayesian networks the degree of connectivity is closely related to the computational complexity of using the network, both space and time complexity. Hence, our encoding scheme generates a preference for more efficient networks. That is, since the encoding length of the model is included in our evaluation of description length, we are enforcing a preference for networks that require the storage of fewer probability parameters and on which exact algorithms are more efficient.

**Encoding the Data Using the Model**   The task is to learn the joint distribution of a collection of random variables $\vec{X} = \{X_1, \ldots, X_n\}$. Each variable $X_i$ has an associated collection of values $\{x_i^1, \ldots, x_i^{s_i}\}$ that it can take on, where the number of values $s_i$ depends on $i$. Every distinct choice of values for all the variables in $\vec{X}$ defines an atomic event in the underlying joint distribution and is assigned a particular probability by that distribution.

We assume that the data points in the raw data are all atomic events. That is, each data point specifies a value for every random variable in $\vec{X}$. Furthermore, we assume that the data points are the result of independent random trials. Hence, we would expect, via the central limit theorem, that each particular instantiation of the variables would eventually appear in the database with a relative frequency approximately equal to its probability. These are standard assumptions.

Given a collection of $N$ data points we want to encode, or store, the data as a binary string. There are various ways in which this encoding can be done, but here we are only interested in using the length of the encoding as a metric, via item 2 in the MDL principle, for comparing the merit of candidate Bayesian Networks. Hence, we can limit our attention to *character codes* [CLR89, pp. 337]. With character codes each atomic event is assigned a unique binary string. Each of the data points, which are all atomic events, is converted to its character code, and the $N$ points are represented by the string formed by concatenating these character codes together. To minimize the total length of the encoding we assign shorter codes to events that occur more frequently. This is the basis for Huffman's encoding scheme. It is well known that Huffman's algorithm yields the shortest encoding of the $N$ data points [LH87].

Say that in the underlying distribution each atomic event $e_i$ has probability $p_i$ and we construct, via some learning scheme, a particular Bayesian network from the raw data. This Bayesian network acts as a model of the underlying distribution and it also assigns a probability, say $q_i$, to every atomic event $e_i$. Of course, in general $q_i$ will not be equal to $p_i$, as the learning scheme cannot guarantee that it will construct a perfectly accurate network. Nevertheless, the aim is for $q_i$ to be close to $p_i$, and the closer it is the more accurate is our model.

The constructed Bayesian network is intended as our best "guess" representation of the underlying distribution. Hence, given that the probabilities $q_i$ determined by the network are our best guess of the true values $p_i$ it makes sense to design our Huffman code using these probabilities. Using the $q_i$ probabilities the Huffman algorithm will assign event $e_i$ a codeword of length approximately $-log_2(q_i)$. If we had the true probabilities $p_i$, the algorithm would have assigned $e_i$ and optimal codeword of length $-log_2(p_i)$ instead. Despite our use of the values $q_i$ in assigning codewords, the raw data will continue to be determined by the true probabilities $p_i$. That is, we still expect that for large $N$ we will have $Np_i$ occurrences of event $e_i$, as $p_i$ is the true probability of $e_i$ occurring. Therefore, when we use the learned Bayesian network to encode the data the length of the string encoding the database will be approximately

$$-N \sum_i p_i \log_2(q_i), \qquad (2)$$

where we are summing over all atomic events. How does this encoding length compare to the encoding length if we had access to the true probabilities $p_i$? An old theorem due originally to Gibbs gives us the answer.

**Theorem 2.1 (Gibbs)** *Let $p_i$ and $q_i$, $i = 1,\ldots,t$, be non-negative real numbers that sum to 1. Then*

$$-\sum_{i=1}^t p_i \log_2(p_i) \le -\sum_{i=1}^t p_i \log_2(q_i),$$

*with equality holding if and only if $\forall i. p_i = q_i$. In the summation we take $0 \log_2(0)$ to be 0.*

In other words, this theorem shows that the encoding using the estimated probabilities $q_i$ will be longer than the encoding using the true probabilities $p_i$. It also says that the true probabilities achieve the minimal encoding length possible.

The MDL principle says that we must choose a network that minimizes the sum of its own encoding length, which depends on the complexity of the network, and the encoding length of the data given the model, which depends on the closeness of the probabilities $q_i$ determined by the network to the true probabilities $p_i$, i.e., on the accuracy of the model.

We could use Equation 2 directly to evaluate the the encoding length of the data given the model. However, the equation involves a summation over all the atomic events, and the number of atomic events is exponential in the number of variables. Instead of trying to use Equation 2 directly we investigate the relationship between encoding length and network topology. Let the underlying joint distribution over the variables $\vec{X} = \{X_1, \ldots, X_n\}$ be $P$. Any Bayesian network model will also define a joint distribution $Q$ over these variables. We can express $Q$ as [Pea88]:

$$Q(\vec{X}) = P(X_1 \mid F_{X_1})P(X_2 \mid F_{X_2})\ldots P(X_n \mid F_{X_n}), \qquad (3)$$

where $F_{X_i}$ is the, possibly empty, set of parents of $X_i$ in the network. Note that $P$ appears on the right hand side instead of $Q$. We obtain the conditional probability parameters on the right from frequency counts taken over the data points. By the law of large numbers we would expect that these frequency counts will be close to the true probabilities over $P$.[3]

We can now prove the following new result that is the basis for our new localized description length computations:

**Theorem 2.2** *The encoding length of the data (Equation 2) can be expressed as:*

$$-N \sum_{i=1}^n W(X_i, F_{X_i}) + N \sum_{i=1}^n [-\sum_{X_i} P(X_i) log_2(P(X_i))] \qquad (4)$$

*where the second sum is taken over all possible instantiations of $X_i$. The term $W(X_i, F_{X_i})$ given by*

$$W(X_i, F_{X_i}) = \sum_{X_i, F_{X_i}} P(X_i, F_{X_i}) \log_2 \frac{P(X_i, F_{X_i})}{P(X_i)P(F_{X_i})} \qquad (5)$$

*where the sum is taken over all possible instantiations of $X_i$ and its parents $F_{X_i}$, and we take $W(X_i, F_{X_i}) = 0$ if $F_{X_i} = \emptyset$. The proof of this, and all other theorems, is presented in our full report [LB93a].*

Given some collection of raw data, the last term in Equation 4 is independent of the structure of the network. Furthermore, the weight measure, the first term in Equation 4, can be calculated locally.

## 3  Localization of the Description Length

To make use of the MDL principle, we need to evaluate the total description length (item 1 + item 2) given a Bayesian network. Adding Equation 1 and 4, the total description length is:

$$\sum_{i=1}^n [k_i log_2(n) + (s_i - 1)(\prod_{j \in F_{X_i}} s_j)d] - N \sum_{i=1}^n W(X_i, F_{X_i})$$
$$+ N \sum_{i=1}^n [-\sum_{X_i} P(X_i) log_2(P(X_i))]$$
$$= \sum_{i=1}^n [[k_i log_2(n) + (s_i - 1)(\prod_{j \in F_{X_i}} s_j)d] - NW(X_i, F_{X_i})]$$
$$+ N \sum_{i=1}^n [-\sum_{X_i} P(X_i) log_2(P(X_i))] \qquad (6)$$

The last term in Equation 6 remains constant for a fixed collection of raw data. Therefore, the first term is sufficient to compare the total description lengths of alternative candidate Bayesian networks.

---

[3] It might not be the case that $P$ is equal to this decomposition. The approximation introduced by our network model is precisely the assumption of such a decomposition.

**Definition 3.1** The *node description length* $DL_i$ for the node $X_i$, with respect to its parents $F_{X_i}$, is defined as:

$$DL_i = k_i log_2(n) + (s_i - 1)(\prod_{j \in F_{X_i}} s_j)d - NW(X_i, F_{X_i}) \quad (7)$$

**Definition 3.2** The *relative total description length* for a Bayesian network, defined as the summation of the node description length of every node in the network, is given by:

$$\sum_{i=1}^{n} DL_i \quad (8)$$

As a result, the relative total description length is exactly equivalent to the first term in Equation 6, and thus is sufficient for comparing candidate networks. Moreover, it can be calculated locally since each $DL_i$ depends only on the set of parent nodes for a given node $X_i$.

**Definition 3.3** Given a collection of raw data, an *optimal Bayesian network* is a Bayesian network for which the total description length is minimum.

Clearly, one or more optimal Bayesian networks must exist for any collection of raw data. Furthermore, we have the following result.

**Theorem 3.4** *Given a collection of raw data, the relative total description length of an optimal Bayesian network is minimum. Also, for a given node $X_i$ in an optimal Bayesian network, $DL_i$ is minimum among those parent sets creating no cycle and not making the network disconnected. That is, we cannot reduce $DL_i$ by modifying the network to change $X_i$'s parents.*

This theorem says that in an optimal network no single node can be locally improved. It is possible, however, that a non-optimal network could also possess this property. In such a case the parent sets of a number of nodes would have to be altered simultaneously in order to reduce its description length.

## 4 Incorporating Partial Domain Knowledge

Although we might not know the underlying joint distribution governing the behavior of the domain variables, we could possibly have other, partial, information about the domain. In particular, our new system can consider two types of domain knowledge: *direct causation specifications* and *partial ordering specifications*.

By direct causation information we mean information of the form "$X_i$ is a direct cause of $X_j$". That is, we might know of a direct causal link between two variables, even if we do not know the causal relationships between the other variables. This kind of information might be provided by, e.g., domain experts, and we can use it when generating the network model. In particular, we can require that in the learned model $X_i$ be one of $X_j$'s parents, thus ensuring that the model validates the direct causation. More generally, the domain experts might be able to construct a skeleton of the network, involving some, but not all, of the variables. The arcs in the skeleton can be specified as direct causation specifications to our system, which will then proceed to fill in the skeleton placing the remaining variables in appropriate positions.

Partial ordering information, on the other hand, specifies ordering relationships between two nodes. Such information might, for example, come from knowledge about the temporal evolution of events in our domain. For instance, if we know that $X_i$ occurs before $X_j$, the network model should not contain a path from $X_j$ to $X_i$ as no causal influence should exist in that direction. Note that a total ordering among the variables, as required by Cooper and Herskovits [CH91], is just a special case of our partial ordering specifications.

Subject to the condition that the direct causation and partial ordering specifications not entail any transitivity violations (e.g., we cannot have a circular sequence of direct causations as input to the system), our system can ensure that the constructed network validates these specifications. Furthermore, information of this sort can in fact lead to increased efficiency: it will constrain our search for an appropriate network model.

To incorporate this information, we define a *constrained Bayesian network* as follows:

**Definition 4.1** A *constrained Bayesian network* is an ordinary Bayesian network whose topology includes all the arcs specified by the direct causation specifications and does not violate any partial ordering specifications.

It can be shown that Theorem 3.4 still holds, with the obvious modifications, if we consider constrained Bayesian networks instead of ordinary networks.

## 5 Searching for the Best Constrained Network

Although our expression for the relative total description length allows us to evaluate the relative merit of candidate network models, we cannot consider all possible networks: there are simply too many of them (an exponential number in fact). Hence, to apply the MDL principle we must engage in a heuristic search that tries to find a good (i.e., low description length), but not necessarily optimal, network model.

In this section we describe our search algorithm which attempts to find a good network by building one up arc by arc. The first step is to rank the possible arcs so that "better" arcs can be added into the candidat-networks before others. The arcs are ranked by calculating the node description length for $X_j$ given the arc $X_i \rightarrow X_j$, $i \neq j$, using Equation 7 and treating $X_i$ as the single parent. This node description length is assigned as the "description length" of arc $X_i \rightarrow X_j$. A list of arcs PAIRS is created sorted so that the first arc on PAIRS has lowest description length. PAIRS will contain all arcs except for those violating the direct causation or partial ordering specifications. Looking at Equation 7 we can see that if $X_i$ and $X_j$ are highly correlated (as measured by $W(X_j, X_i)$, Equation 5) the description length will be lower, and an arc between them will be one of the first that we will try to add to the candidate networks.

Search is performed by a best-first algorithm that maintains OPEN and CLOSED lists each containing search elements. The individual search elements have two components $\langle G, L \rangle$: a candidate network $G$, and an arc $L$ which could be added to the candidate network without causing a cycle or violating the partial ordering and direct causation specifications. OPEN is ordered by heuristic value, which is calculated as the relative total description length (Equation 8) of the element's network, plus the description length of the element's arc (calculated during the construction of PAIRS). Therefore, the lower the heuristic value, the shorter the encoding length. Initially, we construct a network $G_{\text{init}}$ containing only those arcs included in the direct causation specifications. Then, the initial OPEN list is generated by generating all of the search elements $\langle G_{\text{init}}, L \rangle$ for all arcs $L \in$ PAIRS. Best-first search is then executed with the search element at the front of OPEN expanded as follows.

1. Remove the first element from OPEN and copy it onto CLOSED. Let the element's network be $G_{\text{old}}$ and the element's arc be $L$.

2. Invoke the ARC-ABSORPTION procedure on $G_{\text{old}}$ and $L$ to obtain a new network $G_{\text{new}}$ with the arc $L$ added. The ARC-ABSORPTION procedure, described below, might also reverse the direction of some other arcs in $G_{\text{old}}$.

3. Next we make a new search element consisting of $G_{\text{new}}$ and the first arc from PAIRS that appears after the old arc $L$ which could be added to $G_{\text{new}}$ without generating a cycle or violating a partial ordering specification. This new element is placed on OPEN in the correct order according to the heuristic function.

4. Finally, we make another new search element consisting of $G_{\text{old}}$ and the first arc from PAIRS that appears after $L$ which could be added to $G_{\text{old}}$ without generating a cycle or violating a partial ordering specification. Again, this element is placed on OPEN in the correct order.

Now we describe the ARC-ABSORPTION procedure which finds a locally optimal way to insert a new arc into an existing network. To minimize the description length of the resulting network, the procedure might also decide to reverse the direction of some of the old arcs.

Input  : A network $G_{\text{old}}$.
       : An arc $(X_i \rightarrow X_j)$ to be added.
Output : A new network $G_{\text{new}}$ with the arc added and some other arcs possibly reversed.

1. Create a new network by adding the arc $(X_i \rightarrow X_j)$ to $G_{\text{old}}$. In the new network we then search locally to determine if we can decrease the relative total description length by reversing the direction of some of the arcs. This is accomplished via the following steps.

2. Determine the optimal directionality of the arcs attached directly to $X_j$ by examining which directions minimize the relative total description length. Some of these arcs may be reversed by this process.[4] Furthermore, we do not consider the reversal of any arcs that would result in the violation of the direct causation or partial ordering specifications.

3. If the direction of an existing arc is reversed then perform the above directionality determination step on the other node affected.

The search procedure is mainly composed of the ARC-ABSORPTION procedure, a cycle checking routine, and a partial order checking routine. The complexity of cycle checking and partial order checking are $O(n)$ and $O(n^2)$ respectively, where $n$ is the number of nodes. We have found that the search can arrive at a very reasonable network model if provided with a resource bound of $O(n^2)$ search elements expansions. Under this resource bound, we have found that in practice the overall complexity of the search mechanism remains polynomial in the number of nodes $n$.

We can further observe that when the amount of domain information increases, the search time to find a good network model decreases. This arises from the fact that such information places constraints on the space of allowable models making search easier. For example, if a total ordering among the nodes in the domain is given, the search time will be reduced by a factor of $O(n^2)$: there is no need to perform the cycle or partial order checking, and the arc reversal step in ARC-ABSORPTION is no longer needed.

## 6 Experiments

Following [CH91] we test our approach by constructing an original network and using Henrion's logical sampling technique [Hen87] to generate a collection of raw data. We then apply our learning mechanism to the raw data to obtain a learned network. By comparing this network with the original we can determine the performance of our system.

---

[4]Note that it is sufficient to compute the node description length (Equation 7) of those nodes whose parents have been changed. The relative total description length (Equation 8) of the whole network need not be computed.

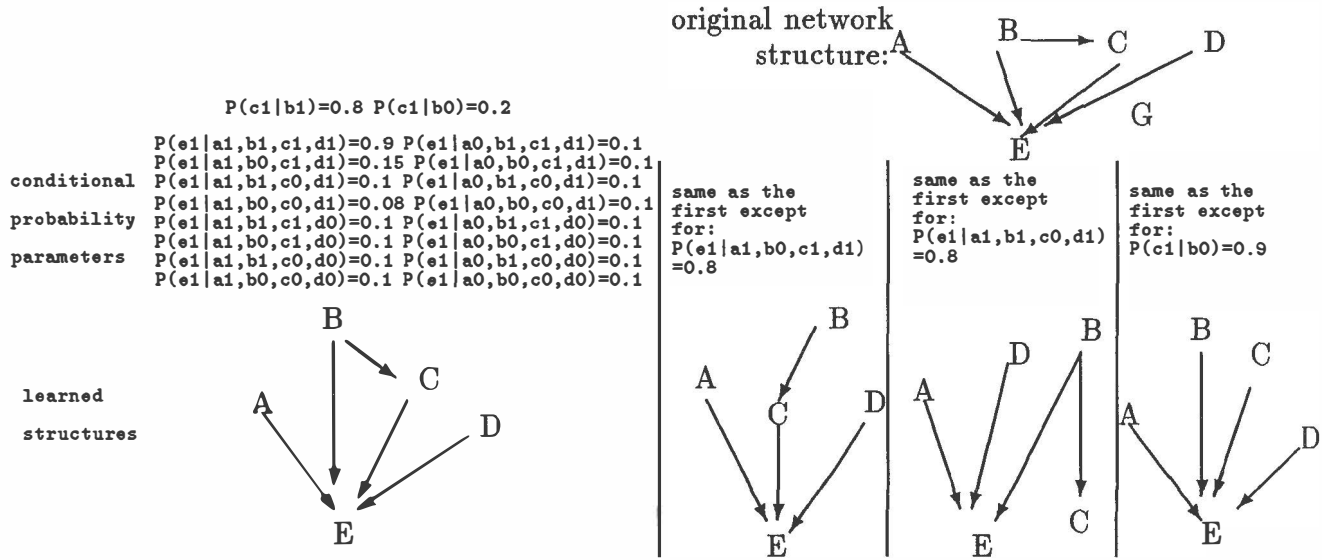

Figure 1: The Quality of Learned Networks

In the first set of experiments, the original Bayesian network $G$ consisted of 5 nodes and 5 arcs. We varied the conditional probability parameters during the process of generating the raw data obtaining four different sets of data. Exhaustive searching, instead of heuristic searching, was then carried out to find the network with minimum total description length for each of these sets of raw data resulting in different learned structures in each case. The experiment demonstrates that our algorithm does in fact yield a tradeoff between accuracy and complexity of the learned structures: in all cases where the original network was not recovered a simpler network was learned. The type of structure learned depends on the parameters, as each set of parameters, in conjunction with the structure, defines a different probability distribution. Some of these distributions can be accurately modeled with simpler structures. In the first case, the distribution defined by the parameters did not have a simpler model of sufficient accuracy, but in the other cases it did. We have also developed measures of the absolute accuracy of the learned structures (see [LB93b] for a full description) that indicate in all cases that the learned structure was very accurate even though it might possess a different topology.

The second experiment consisted of learning a Bayesian network with a fairly large number of variables (37 nodes and 46 arcs). This network was derived from a real-world application in medical diagnosis [BSCC89] and is known as the ALARM network (see [LB93b] for a diagram of this network). After applying our heuristic search algorithm, we found that the learned network is almost identical to the original structure with the exception of one different arc and one missing arc. One characteristic of our heuristic search algorithm is that we did not require a user supplied ordering of variables (cf. Cooper and Herskovits [CH91]). This experiment demonstrates the feasibility of our approach for recovering networks of practical size.

Besides being able to use extra domain information our new search mechanism is faster and more accurate than the mechanism first reported in [LB93b] which was developed without the local measure of description length. To investigate how our search mechanism behaves when domain information is supplied, we conducted some further experiments. Using the same set of raw data derived from the ALARM model in conjunction with varying amounts of domain information, we applied our learning algorithm and recorded the search time required to obtain an accurate network model. The following two tables depict the relative time required by the search algorithm when varying amounts of direct causation and partial orderings specifications are made available. In general, the search time decreases as the amount of causal information increases.

|  | no partial ordering | 10 partial orderings | 20 partial orderings | total ordering |
|---|---|---|---|---|
| time | 100% | 84% | 60% | 20% |

|  | no direct causal specification | 10 direct causal specifications | 20 direct causal specifications |
|---|---|---|---|
| time | 100% | 74% | 25% |

## 7 Refinement of Existent Networks

Besides the advantages outlined above our new local computation of description length also allows for the possibility of refining an existing network by modifying some local part of it. Refinement is based on the following theorem.

**Theorem 7.1** *Let $\vec{X} = \{X_1, X_2, \ldots, X_n\}$ be the nodes in an existent Bayesian network, $X'$ be a subset of $\vec{X}$, and $DL_{X'}$ be the total node description lengths of all the nodes in $X'$ (i.e., $DL_{X'} = \sum_{X_i \in X'} DL_i$). Suppose we find a new set of parents for every node in $X'$ that does not create any cycles or make the network disconnected. Let the new total node description lengths of all the nodes in $X'$ be $DL_{newX'}$. Then we can construct a new network in which the parents of the nodes in $X'$ are replaced by their new parent sets, such that the new network will have lower total description length if $DL_{newX'} < DL_{X'}$.*

This theorem provides a means to improve a Bayesian network without evaluating the total description length of the whole Bayesian network, a potentially expensive task if the network is large. We can isolate a subset of nodes and try to improve that collection locally, ignoring the rest of the network. Algorithms for performing such a refinement, based on this theorem, have been developed and experiments are being performed. We hope to report on this work in the near future [LB93a].